\documentclass[onecolumn,11pt]{IEEEtran}
 \newtheorem{thm}{Theorem}[section]
 \newtheorem{cor}[thm]{Corollary}
 \newtheorem{prop}[thm]{Proposition}
 \newtheorem{defn}[thm]{Definition}
 \newtheorem{alg}[thm]{Algorithm}
 \newcommand{\abs}[1]{\left\vert#1\right\vert}

\begin{document}
%
\title{The Minimal Cost Algorithm for off-line  Diagnosability
of Discrete Event Systems}
\author{Zhujun Fan\\
{\footnotesize   Department of Mathematics, Zhongshan University,
Guangzhou 510275, China}\\
}

\date{}
\markboth{The Minimal Cost Algorithm for off-line  Diagnosability
of Discrete Event Systems }{Shell \MakeLowercase{\textit{et al.}}:
Bare Demo of IEEEtran.cls for Journals}
\maketitle

\begin{abstract}
The failure diagnosis for {\it discrete event systems} (DESs) has
been given considerable attention in recent years. Both on-line
and off-line diagnostics in the framework of DESs was first
considered by Lin Feng in 1994, and particularly an algorithm for
diagnosability of DESs was presented. Motivated by some existing
problems to be overcome in previous work, in this paper, we
investigate the minimal cost algorithm for diagnosability of DESs.
 More specifically: (i) we give a generic method for judging a system's off-line
diagnosability, and the complexity of this algorithm is
polynomial-time; (ii) and in particular, we present an algorithm
of how to search for the minimal set in all observable event sets,
whereas the previous algorithm may find {\it non-minimal} one.
\end{abstract}

\IEEEpeerreviewmaketitle

\section{Introduction}
\PARstart{W}{ith} man-made systems becoming more and more complex,
detecting and locating component failure is not a simple task.
Therefore, there is a strong need for a systematic study of
diagnostic problems and diagnosability issues [30]. As an important
kind of  man-made systems, discrete event system (DES) is a
dynamical system whose state space is discrete and whose states can
only change as a result of asynchronous occurrence of instantaneous
events over time [2]. Up to now, DESs have been successfully applied
to provide a formal treatment of many technological and engineering
systems [3, 5, 16]. Naturally, the diagnosability of DESs is of
theoretical and practical importance.

\par Actually, diagnosability of DESs has received extensive attention in
recent years ( for example, [1, 4, 6-9, 11-15, 17-32]).
Especially, in [15], the definitions of ``off-line''
diagnosability and ``on-line'' diagnosability were introduced, and
both ``off-line'' diagnostic algorithm and ``on-line'' diagnostic
algorithm were significantly established in the framework of DESs.
However, the algorithms presented in [15] have some shortcomings:
1) the computational complexity of ``off-line'' diagnostic
algorithm is exponential in general; 2) and ``off-line''
diagnostic algorithm could not find the minimal one in observable
events sets (OESs), and the algorithm of how to inspect an
automaton being diagnosable was not yet given. Motivated by these
issues, our goal in this paper is to solve these problems.

\par The remainder of the paper is organized as follows. In Section
~\ref{S:Prepare}, we first introduce a general framework of
diagnosability of DESs, and then explain the lost of Algorithm 1
in ~\cite{Lin94}. In Section ~\ref{S:OFD}, the definition of
``off-line'' diagnostics is first provided, and we then present a
polynomial-time algorithm to realize it. In Section
~\ref{S:NAOES}, we demonstrate the principle of finding the
minimal set in an OES, and particularly, present our new algorithm
to realize it; Section ~\ref{S:EXP} provides two examples to
illustrate these algorithms in Sections ~\ref{S:OFD} and
~\ref{S:NAOES}. Finally some remarks are made in Section VI to
conclude the paper.

\section{Preliminaries}\label{S:Prepare}
\subsection{A general
framework for automata and diagnostics}
\subsubsection{$DFAs$} A
deterministic finite automaton ($DFA$) can be formally defined as
a 5-tuple $(Q, \Sigma, \delta, q_{0}, F)$, where $Q$ is a finite
set of states, $\Sigma$ is the input alphabet,
$\delta:Q\times\Sigma\rightarrow Q$ is the transition function,
$q_{0}\in Q$ is the starting state, and $F\subseteq Q$ is a set of
 accepting states. Operation of the $DFA$ begins at
$q_{0}$, and movement from state to state is governed by the
transition function $\delta$. $\delta$ must be defined for every
possible state in $Q$ and every possible symbol in $\Sigma$.

 A $DFA$ can be represented visually as a directed
graph. Circular vertices denote states, and the set of directed
edges, labeled by symbols in $\Sigma$, denotes $\delta$. The
transition function takes the first symbol of the input string,
and after the transition this first symbol is removed. If the
input string is $\epsilon$ (the empty string), then the operation
of the $DFA$ is halted. If the final state when the $DFA$ halts is
in $F$, then the $DFA$ can be said to have accepted the input
string it was originally given. The starting state $q_{0}$ is
usually denoted by an arrow pointing to it that points from no
other vertex. States in $F$ are usually denoted by double circles.

$DFAs$ recognizes regular languages, and can be used to test
whether any string is in the language it recognizes. As it is
known, DFAs have been used to model DESs [2]. In the following, we
use $DFA$ to represent a DES.

\subsubsection {Diagnosability of Discrete Event Systems} We model the
system to be diagnosed as a pair $G=(M,\Sigma_{c})$. The first
component $M$ denotes a nondeterministic Mealy automaton:
\[M=(\Sigma,Q,Y, \delta,h)\]
where $\Sigma$ is the set of finite events; $Q$ is the set of finite
states; $Y$ is the output alphabet space; $\delta:\Sigma\times
Q\rightarrow 2^{Q}$ is the state transition function.
$\delta(\sigma,q)$ gives the set of possible next states if $\sigma$
occurs at $q$; and $h:\Sigma\times Q\rightarrow Y$ is the output
function, $h(\sigma,q)$ is the observed output when $\sigma$ occurs
at $q$. The second component $\Sigma_{c} \subseteq \Sigma$ is the
set of controllable events, where the controllability of events is
interpreted in a strong sense: a controllable event can be made to
occur if physically possible.
\par
States of the system describe the conditions of its components.
Therefore, to diagnose a failure is to identify which state or set
of states the system belongs to. Thus, depending on the requirements
on diagnostics, we partition the state space $Q$ into disjoint
subset (cells) and denote the desired partition by $T$. The state in
the same cell are viewed as equivalent as far as failures under
consideration are concerned. The model is rather general since we do
not put any restrictions on $T$.
\subsubsection{Some notations}
\par For convenience, we give some notations. Let
\[M=\{\sigma_{1},\sigma_{2},\ldots,\sigma_{n}\},\]
where $\sigma_{i}, i=1,\ldots,n$, are the observed events, and the
cost of $M$ is denoted by
\[C(M)=\{c(\sigma_{1}),c(\sigma_{2}),\ldots,c(\sigma_{n})\}.\]
where $c(\sigma_{i})$ means the cost of observes event
$\sigma_{i}, i=1,\ldots,n$, respectively. Without loss of
generality, we suppose that
\[c(\sigma_{1})\geq c(\sigma_{2})\geq\ldots\geq c(\sigma_{n}).\]
Let $\Sigma_{o}\in OES$. We denote
$C(\Sigma_{o})=\sum\limits_{\sigma_{i}\in \Sigma_{o}}
 c(\sigma_{i})$, which presents the cost of $\Sigma_{o}$, and
\[minL(\Sigma_{o})=min\{i: \sigma_{i}\in \Sigma_{o}\}.\]
\par An important problem is how to find the smallest observable event
set that makes $G$ diagnosable for a given partition $T$. In order
to solve this problem, we define the set of all observable event
sets (OESs) that ensure the
diagnosability of the system as:\\
 $OES(T)$ = \{$\Sigma_{o} \subseteq\Sigma$: $G$ is
diagnosable with respect to $\Sigma_{o}$ and $T$\}.
\subsection{Lost of Algorithm 1 of~\cite{Lin94}}
\subsubsection{Algorithm 1 of~\cite{Lin94}} In order to remove
events one by one in the given order until the diagnosability of
the system is no longer ensured, Algorithm~\ref{A:MINOES}
(Fig.~\ref{F:MINOES}) was presented in [15].
\begin{figure}
\fbox{\parbox{9cm}{
 {\small
 \begin{alg}(minOES)\label{A:MINOES}
\begin{quote} {\bf Input:}
 Read $G=(M, \Sigma_{C})$, $M=(\Sigma,Q,Y,\delta,h)$,$T$;
\begin{quote}{\bf Initialization:}$minOES$ := $\Sigma$;
\end{quote}
\begin{quote}{\bf Removal:}
{\bf For} $i = 1$ to $n$ {\bf do}\\
{\bf begin} $minOES := minOES \setminus \{\sigma_{i}\}$;\\
{\bf if} $G$ is not diagnosable with respect to $minOES$ an $T$ {\bf
then}
\begin{quote}$minOES := minOES\bigcup\{\sigma_{i}\};$
\end{quote}
{\bf end};
\end{quote}
\end{quote}
\begin{quote}{\bf Output:}Return $minOES$;
\end{quote}
\end{alg}
}}} \caption{Algorithm 1 of~\cite{Lin94}.} \label{F:MINOES}
\end{figure}
\par
However, Algorithm~\ref{A:MINOES} has some shortcomings, we will
illustrated them in next subsection.

\subsubsection{Lost Example of  Algorithm 1 of~\cite{Lin94}}\label{S:LEA} In
fact, the ``off-line'' diagnostic algorithm
(Algorithm~\ref{A:MINOES}) could not find the minimal one in
observable events sets. For example, let
\[M=\{\sigma_{1},\sigma_{2},\sigma_{3},\sigma_{4},\sigma_{5}\},\]
and the cost of $M$ is
\[C(M)=\{13, 9, 7, 5, 2\}.\]
\par $OES(T) = \{E_{1}\}\cup
\{E_{2}\}\cup\{\Sigma\subseteq M: E_{1}\subseteq\Sigma,or,
E_{2}\subseteq\Sigma\}$, where $E_{1} =\{\sigma_{2},
\sigma_{5}\}$, and $E_{2} = \{\sigma_{3}, \sigma_{4},
\sigma_{5}\}.$
\par $G$ is diagnosable with respect to given $T$ and an element of
 $OES(T)$.
\par If we use Algorithm~\ref{A:MINOES}, we can get the $minOES =
E_{2}$, the cost of $E_{2}$ is $7+5+2=14$. However, the cost of
$E_{1}$ is $9+2=11$, which is less than the cost of $E_{2}$.
Therefore, the $minOES$ is not the minimal cost of $OES(T)$.

\section{Off-line Diagnostics}\label{S:OFD}
Off-line diagnostics means that diagnosis is performed when the
system is not in normal operation [15]. For example, what a
mechanic does to an automobile in a repair shop can be viewed as
off-line diagnostics. In order to perform off-line diagnostics,
one can ``open'' the system, access the inside, do various tests,
and measure responses that may not be available from the system
outputs. In fact, during off-line diagnostics, the system is not
actually in operation. Therefore, the failure status of system
components will not change, unless such changes are made in
purpose. So tests can be designed with great flexibility and the
order of testing is not critical as far as diagnosability is
concerned.

\subsection{Off-line Diagnostics ~\cite{Lin94}}
For off-line diagnostics, we specialize the model introduced in the
previous section by assuming that the output are events observed.
That is, $Y = \Sigma_{o}$, where $\Sigma_{o}\subseteq \Sigma$ is the
set of observable events and the output map $h:\Sigma\times
Q\rightarrow\Sigma_{o}$ is a projection defined as:
\[h(\sigma,q)=\left\lbrace
\begin{array}{c c}
\sigma &  {\rm if}\hskip 2mm  \sigma\in\Sigma_{o},\\
\epsilon & {\rm otherwise},
\end{array}
\right.\]
where $\epsilon$ is the empty string.
\par
As it was discussed before, in off-line diagnostics all events are
assumed to be controllable. Therefore, $\Sigma_{c} =\Sigma$. Since
the failure states of system components will not change, information
derived from all the test outputs are updated and relevant.
\par
During off-line diagnostics, if an event $\sigma\in\Sigma_{o}$ is
observed, then the possible state of the system is:
\begin{equation}\label{E:QC}
Q(\sigma)=\{q\in Q:(\exists q'\in Q)\delta(\sigma,q')=q\}.
\end{equation}
Hence, we know every state of the system is in either $Q(\sigma)$ or
$Q-Q(\sigma)$ after observing $\sigma$. That is, each observable
event partitions the state space into:
\begin{equation}\label{E:TQC}
T_{\sigma}=\{Q(\sigma),Q-Q(\sigma)\}.
\end{equation}
\par
Since there is not restriction on the tests performed in off-line
diagnostics, we can observe all observable events that are
physically possible and then determine which states the system is
in. If this information is sufficient for us to determine which
component is broken (i.e., which cell of $T$ the system is in), then
we say the system is off-line diagnosable. Formally:
\begin{defn}\label{D:OFD}
$G$ is said to be off-line diagnosable with respect to $T$ if
\begin{equation}\label{E:OFD}
\bigwedge\limits_{\sigma\in \Sigma_{o}} T_{\sigma} \leq T
\end{equation}
 where
$\wedge$ denotes conjunction and $\leq$ means ``is finer than''.
\end{defn}
\par
Clearly, diagnosability depends on both the observable event set
$\Sigma_{o}$ and the desired partition $T$.

\subsection{ An algorithm for off-line diagnosability}\label{S:AOFD}
In Section ~\ref{S:OFD}, we have introduced the definition of
``off-line'' diagnostics (see Definition~\ref{D:OFD} and equation
(\ref{E:OFD}) in Section ~\ref{S:OFD}). In equation (\ref{E:OFD}),
the right part $T$ is given by the system. Now we must first
calculate the left part $\bigwedge\limits_{\sigma\in \Sigma_{o}}
T_{\sigma}$, where $T_{\sigma}$ is given in equation (\ref{E:TQC})
of Section ~\ref{S:OFD} and $Q(\sigma)$ is given in equation
(\ref{E:QC}) of Section ~\ref{S:OFD}. From these two equations,
given an element $\sigma$, for every element of $Q$, it must be in
$Q(\sigma)$ or not in $Q(\sigma)$ (in $Q \setminus Q(\sigma)$). So
we can use one bit to identify every element of $Q$ in $Q(\sigma)$
or not in $Q(\sigma)$ (i.e, 1  for elements in $Q(\sigma)$ and 0 for
elements not in $Q(\sigma)$). Now we give algorithms to realize
them.
\subsubsection{Algorithm}
\par Algorithm~\ref{A:QC} (Fig.~\ref{F:QC}) gives whether a state q in $Q(\sigma)$ or not, that
is $\delta(\sigma,Q)=q$ or $\delta(\sigma,Q)\neq q$.
\par Algorithm~\ref{A:TQC} (Fig.~\ref{F:TQC}) gives the calculation
of $\bigwedge\limits_{\sigma\in \Sigma_{o}} T_{\sigma}$.
\begin{figure}
\fbox{\parbox{8cm}{
 {\small
\begin{alg}(QC).\label{A:QC}
\begin{quote} {\bf Input:}
$\delta, \sigma, Q, q$;
\begin{quote}{\bf Initialization:}
Set $m=\abs{Q},QC:=False$;
\end{quote}
\begin{quote}{\bf Judge:}
{\bf for} $i = 1$ to $m$ {\bf do}
\begin{quote}{\bf if} $\delta(\sigma,q_{i}) == q$ {\bf then}
\begin{quote} $QC:=True$,{\bf break};\end{quote}
\end{quote}
\end{quote}
\end{quote}
\begin{quote}{\bf Output:}
Return QC;
\end{quote}
\end{alg}
}}} \caption{Algorithm:whether a state q in $Q(\sigma)$ or not, that
is $\delta(\sigma,Q)=q$ or $\delta(\sigma,Q)\neq q$.} \label{F:QC}
\end{figure}

\begin{figure}
\fbox{\parbox{9cm}{
 {\small
\begin{alg}(TQC).\label{A:TQC}
\begin{quote} {\bf Input:}
$\delta, \Sigma_{o}, Q$;
\begin{quote}{\bf Initialization:}
Set $m=\abs{Q}, n=\abs{\Sigma_{o}}, s_{j}=0(j=1..m)$
\end{quote}
\begin{quote}{\bf Intersection:}
{\bf for} $i = 1$ to $n$ {\bf do}\\
{\bf for} $j = 1$ to $m$ {\bf do}
\begin{quote}{\bf if} $\delta(\sigma_{i},Q) = q_{j}$(Algorithm ~\ref{A:QC}) {\bf
then}
\begin{quote}$s_{j} |= (1 << (i-1))$;
\end{quote}
\end{quote}
\end{quote}
\end{quote}
\begin{quote}{\bf Output:}
Return $s_{j}(j=1..m)$;
\end{quote}
\end{alg}
}}} \caption{Algorithm:$\bigwedge\limits_{\sigma\in \Sigma_{o}}
T_{\sigma}$.} \label{F:TQC}
\end{figure}

\par In Algorithm ~\ref{A:TQC}, all the elements of $F$ $(F\in\bigwedge
\limits_{\sigma\in \Sigma_{o}} T_{\sigma}$) have the same value
$s_{j}$, since they have the same operation in Algorithm
~\ref{A:TQC}. And $\bigwedge\limits_{\sigma\in \Sigma_{o}}
T_{\sigma} \leq T$ means that every element of
$\bigwedge\limits_{\sigma\in \Sigma_{o}} T_{\sigma}$ is the subset
of $G$ $(G\in T)$. The reverse proposition means that there exists
an element of $\bigwedge\limits_{\sigma\in \Sigma_{o}}
T_{\sigma}$, not all of its elements are the elements of $G(G\in
T)$. From this, we have Algorithm ~\ref{A:OFD} (Fig.~\ref{F:OFD}).
\begin{figure}
\fbox{\parbox{8cm}{
 {\small
 \begin{alg}(OFD).\label{A:OFD}
\begin{quote} {\bf Input:}
$\delta, \Sigma_{o}, Q, T$;
\begin{quote} {\bf Initialization:}
Set $OFD := True$;
\end{quote}
\begin{quote} {\bf Diagnosing:}
Get $s_{j}(j=1..m)$ from Algorithm ~\ref{A:TQC};\\
Applied Quicksort Algorithm to $s_{j}(j=1..m)$;\\
{\bf for} $j = 1$ to $m-1$ {\bf do}
\begin{quote} {\bf if} $(s_{j} == s_{j+1})$ {\bf then}
\begin{quote} {\bf begin} Find $T_{i}\in T$, s.t.$\sigma_{j} \in
T_{i}$;\\
{\bf if} $\sigma_{j+1} \overline{\in} T_{i}$ {\bf then}
\begin{quote} $OFD := False$;\end{quote}
{\bf end};
\end{quote}
\end{quote}
\end{quote}
\end{quote}
\begin{quote} {\bf Output:}
Return $OFD$;
\end{quote}
\end{alg}
}}} \caption{Algorithm:$\bigwedge\limits_{\sigma\in \Sigma_{o}}
T_{\sigma} \leq T$.} \label{F:OFD}
\end{figure}
\subsubsection{Algorithm Complexity}
\par In Algorithm ~\ref{A:QC}, in ``Judge'' recycle, the bad time is $m$. So the
time complexity of Algorithm ~\ref{A:QC} is $O(m)$.
\par In Algorithm ~\ref{A:TQC}, in ``Intersection'' recycle, it has two loops, the
complexity of first line is $O(n)$; the complexity of second line
is $O(m)$. In third line, it calls the Algorithm ~\ref{A:QC}, so
the bad time is $O(m)$;  and then the total complexity in
``Intersection'' recycle is $O(m^{2}n$). Therefore, the time
complexity of Algorithm ~\ref{A:TQC} is $O(m^{2}n$).

\par In Algorithm ~\ref{A:OFD}, in ``Diagnosing'' recycle, it first calls the Algorithm
~\ref{A:TQC},  the time complexity is $O(m^{2}n$); then it calls the
Quicksort Algorithm, the bad time complexity is $O(m^{2}$); for the
other lines, it has one loop, the total complexity is $O(m^{2}$). In
conclusion, the time complexity of Algorithm ~\ref{A:OFD} is
$O(m^{2}n$).

\section{New Algorithm for finding the minimal one in $OESs$}\label{S:NAOES}
\subsection{Finding the minimal one in $OESs$}

 We would like
to find a minimal element in $OES(T)$ as follows.
\begin{prop}\label{p:LIN1}
If $OES(T)$ is not null, then the minimal elements of $OES(T)$
exist, but may not be unique.
\end{prop}
\begin{proof}
The proof of the existence of minimal elements is straightforward.
Since $\Sigma$ is finite,  $2^{\Sigma}$ is a finite set.  Notice
that $\Sigma_{o} \subseteq \Sigma$, therefore, $\Sigma_{o}$ is an
element of $2^{\Sigma}$, and  the elements of $OES(T)$ are finite.
As a result, there exists a minimal element in $OES(T)$ .

The following example shows that the minimal elements of $OES(T)$
may not be unique. Let
\[\begin{array}{r c l}
\Sigma & = & \{\alpha,\beta,\gamma\}\\
Q & = & \{q_{1},q_{2}\}\\
\delta(\alpha,q_{1}) & = &\{q_{2}\}\\
\delta(\beta,q_{2})& = & \{q_{1}\}\\
\delta(\gamma,q_{1})& = & \{q_{2}\}\\
\delta(\sigma,q) & = & \emptyset \texttt{   }  otherwise,
\end{array}\] and
\[T=\{\{q_{1}\},\{q_{2}\}\}.\]
Obviously,  $\{\alpha\}$, $\{\beta\}$ and $\{\gamma\}$ are minimal
elements of $OES(T)$.
\end{proof}
\par From Proposition~\ref{p:LIN1} in Section ~\ref{S:OFD}, we conclude
that we may be able to find more than one set of observable events,
 and each set is minimal in the sense that removing any event from the
set will make the system not diagnosable. Practically, we can find a
cost-effective minimal observable event set by first ordering the
events in terms of the difficulty (and hence cost) in detection.
This directly gives the Algorithm1 of
~\cite{Lin94}(Fig.~\ref{F:MINOES} Algorithm~\ref{A:MINOES}).

\subsection{New Algorithm}
\par From the Example in Section~\ref{S:LEA}, we know that the $minOES$
given by Algorithm~\ref{A:MINOES} is not the minimal cost one.
Therefore, we will modify Algorithm~\ref{A:MINOES} to find the
minimal cost one in this subsection.
\begin{prop}
By Algorithm~\ref{A:MINOES}, we get the $minOES$, whose cost is
$C(minOES)$ and whose minimal label is $minL(minOES)$. If there
exists an $\Sigma_{o}\in OES(T)$, with $C(\Sigma_{o})< C(minOES)$,
then we have
\[minL(\Sigma_{o}) \leq minL(minOES).\]
\end{prop}
\begin{proof} If $minL(\Sigma_{o}) > minL(minOES)$.
Set $L=minL(minOES)$ is the minimal index of set $minOES$. In
Algorithm~\ref{A:MINOES}, when $I = L$, $G$ is diagnosable with
respect to $minOES$ and $T$, and the next step of
Algorithm~\ref{A:MINOES} is not executed. So $\sigma_{L}
\overline{\in} minOES$, and then $L \neq minL(minOES)$.
Consequently,  $minL(\Sigma_{o}) \leq minL(minOES)$.
\end{proof}

Now we present a new algorithm~\ref{A:MMOES} (Fig.~\ref{F:MMOES})
to find the minimal cost one.
\begin{figure}
\fbox{\parbox{8cm}{
 {\small
\begin{alg}(MMOES)\label{A:MMOES}.
\begin{quote} {\bf Input:}
$G=(M, \Sigma_{C})$, $M=(\Sigma,Q,Y, \delta,h)$,T; the order
$\Sigma=\{\sigma_{1},\sigma_{2},\ldots,\sigma_{n}\}$; the cost of
$\Sigma,
C(\Sigma)=\{c(\sigma_{1}),c(\sigma_{2}),\ldots,c(\sigma_{n})\}$;
\begin{quote} {\bf Initialization:}
Get $minOES$ by Algorithm~\ref{A:MINOES};\\
Set $lmS= minL(minOES)$, $cmS= C(minOES)$;\\
Set $H = \{\Sigma_{o}\subseteq\Sigma:minL(\Sigma_{o}) \geq
lmS,C(\Sigma_{o}) < cmS\}$; \\Set $ng = \abs{H}$, and $H =
\{H_{1},H_{2},\ldots,H_{ng}\}$;
\end{quote}
\begin{quote} {\bf Testing diagnosability:}
{\bf for} $i = 1$ to $ng$ {\bf do}
\begin{quote} {\bf begin if} ($G$ is diagnosable with respect to $H_{i}$ an T) AND ($C(H_{i}) < cmS$) {\bf
then}
\begin{quote}$minOES = H_{i}, cmS= C(H_{i})$;\end{quote}
{\bf end};
\end{quote}
\end{quote}
\end{quote}
\begin{quote} {\bf Output:}
Return $minOES$;
\end{quote}\end{alg}
}}} \caption{Algorithm:Modify $MinOES$.} \label{F:MMOES}
\end{figure}
\subsection{Necessary Element}
\begin{defn}\label{d:NE}
(Necessary Element) Suppose  $\Sigma\in OES(T)$. If
$\sigma_{i}\in\Sigma$, but $\Sigma \setminus \{\sigma_{i}\}
\overline{\in} OES(T)$, then we call $\sigma_{i}$ necessary element
with respect to $T$.
\end{defn}
\begin{prop}
If  $\Sigma_{o}\in OES(T)$, and $\Sigma_{o}\subseteq F$, then $F\in
OES(T)$.
\end{prop}
\begin{proof}
Because
\[\bigwedge\limits_{\sigma\in F} T_{\sigma}\leq\bigwedge
\limits_{\sigma\in \Sigma_{o}} T_{\sigma}\leq T,\]  the proposition
holds true.
\end{proof}
\begin{prop}\label{p:NE}
 If $\sigma_{i}$ is a necessary element, then for any $\Sigma_{o}\in OES(T)$,
   $\sigma_{i}\in \Sigma_{o}$.
\end{prop}
\begin{proof}
(proof by contradiction)  If the theorem is not true, there exists
an $\Sigma_{o}\in OES(T)$, with $\sigma_{i}\overline{\in}
\Sigma_{o}$. Therefore,
$\Sigma_{o}\subseteq\Sigma\setminus\{\sigma_{i}\}$. And
$\sigma_{i}$ is a necessary element,
$\Sigma\setminus\{\sigma_{i}\}\overline{\in}OES(T)$. So
$\Sigma_{o} \overline{\in} OES(T)$, which is a contradiction to
assumption. So the proposition is true.
\end{proof}
\begin{defn}\label{d:NES}
(Necessary element set)  $NES(T)$=\{ $\sigma_{i}$ :  $\sigma_{i}$
is necessary element with respect to $T$\}.
\end{defn}
\begin{cor}
For any $\Sigma_{o}\in OES(T)$, $NES(T)\subseteq \Sigma_{o}$.
\end{cor}
\begin{proof}
For any $\Sigma_{o}\in OES(T)$, and any $\sigma_{i}\in NES(T)$,
there exists $\sigma_{i}\in \Sigma_{o}$ (see
Proposition~\ref{p:NE}). So $NES(T)\subseteq \Sigma_{o}$.
\end{proof}

\begin{figure}
\fbox{\parbox{8cm}{
 {\small
 \begin{alg}(NES).\label{A:NES}
\begin{quote} {\bf Input:}
$G=(M, \Sigma_{C})$, $M=(\Sigma,Q,Y,\delta,h)$,T; the order
$\Sigma=\{\sigma_{1},\sigma_{2},\ldots,\sigma_{n}\}$;
\begin{quote} {\bf Initialization:}
Set $NES := \emptyset$;
\end{quote}
\begin{quote} {\bf AddElement:}
{\bf for} $i = 1$ to $n$ {\bf do}
\begin{quote} {\bf begin if} $G$ is not diagnosable with respect to
$\Sigma\setminus\{\sigma_{j}\}$ and $T$ {\bf then}
\begin{quote}$NES := NES\cup\{\sigma_{j}\}$;\end{quote}
{\bf end};
\end{quote}
\end{quote}
\end{quote}
\begin{quote} {\bf Output:}
Return $NES$;
\end{quote}\end{alg}
}}} \caption{Algorithm:NES(T).}
\end{figure}\par We introduce $NES(T)$ to reduce the computing time. We
partition the finite events space $\Sigma$ into two disjoint subsets
$NES(T)$ and $\Sigma\setminus NES(T)$. The set $NES(T)$ must include
all the elements in $OES(T)$. If we get $NES(T)$ first, Algorithm
~\ref{A:MMOES} in this section need only compute in set
$\Sigma\setminus NES(T)$. This may reduce computing complexity.
\subsection{Algorithm Complexity}
Suppose the time (of whether G is not diagnosable with respect to
$minOES$ and $T$) is $T_{G}$, where $T_{G}=O(m^{2}n$).
\par In Algorithm ~\ref{A:MINOES}, in ``Removal'' recycle, the
bad time is $n\times T_{G}$, and the time-complexity of
Algorithm~\ref{A:MINOES} is $O(m^{2}n^{2})$.
\par In Algorithm ~\ref{A:MMOES},in ``Initialization'' recycle, it
first calls Algorithm ~\ref{A:MINOES} to get $minOES$, the bad time
is $O(m^{2}n^{2})$; and then it get set $H$, its a 0-1 pack problem,
the bad time is $O(n\times cmS)$; in ``Testing diagnosability''
recycle, the bad time is $ng\times T_{G}$; therefore the
time-complexity of Algorithm~\ref{A:MMOES} is $O(m^{2}\times n\times
ng)$.
\par In Algorithm~\ref{A:NES}, in ``AddElement'' recycle, the bad time
is $n\times T_{G}$, so the time-complexity of Algorithm~\ref{A:NES}
is $O(m^{2}n^{2})$.
\par Because $\Sigma_{o}\subseteq\Sigma$, $n=\abs{\Sigma}$ in this
section is greater than  $n=\abs{\Sigma_{o}}$ in Section
~\ref{S:AOFD}.
\section{Examples}\label{S:EXP}
\subsection{Example of Algorithm in Section ~\ref{S:AOFD}}
Let us consider the system which is visualized as
Fig.~\ref{F:ATMATA}:

\begin{figure}
\begin{picture}(210,210)(10,20)
\put(110, 110){\makebox(4,2)[l]{\hbox{$q_{0}$}}} \put(40,
40){\makebox(4,2)[l]{\hbox{$q_{3}$}}} \put(40,
180){\makebox(4,2)[l]{\hbox{$q_{1}$}}} \put(155,
40){\makebox(4,2)[l]{\hbox{$q_{4}$}}} \put(155,
180){\makebox(4,2)[l]{\hbox{$q_{2}$}}} \put(100, 110){\vector(1,1)
{68}}\put(100, 110){\vector(-1,-1) {68}} \put(100,
110){\vector(-1,1) {68}}\put(100, 110){\vector(1,-1) {68}}\put(100,
110){\circle*{4}}\put(30, 40){\circle*{4}}\put(30,
180){\circle*{4}}\put(170, 40){\circle*{4}}\put(170,
180){\circle*{4}}\put(180, 190){\circle{26}}\put(180,
30){\circle{26}}\put(20, 190){\circle{26}}\put(20,
30){\circle{26}}\put(110, 120){\circle{26}} \put(10,
210){\makebox(4,2)[l]{\hbox{$\sigma_{1},\sigma_{4}$}}}\put(170,
210){\makebox(4,2)[l]{\hbox{$\sigma_{1},\sigma_{3}$}}}\put(170,
10){\makebox(4,2)[l]{\hbox{$\sigma_{2},\sigma_{4}$}}}\put(10,
10){\makebox(4,2)[l]{\hbox{$\sigma_{2},\sigma_{3}$}}}\put(127,
115){\makebox(4,2)[l]{\hbox{$\sigma_{4}$}}}\put(45,
80){\makebox(4,2)[l]{\hbox{$\sigma_{2},\sigma_{3}$}}}\put(135,
80){\makebox(4,2)[l]{\hbox{$\sigma_{2},\sigma_{4}$}}}\put(45,
135){\makebox(4,2)[l]{\hbox{$\sigma_{1},\sigma_{4}$}}}\put(135,
135){\makebox(4,2)[l]{\hbox{$\sigma_{1},\sigma_{3}$}}}
\end{picture}
\caption{An Automata.} \label{F:ATMATA}
\end{figure}
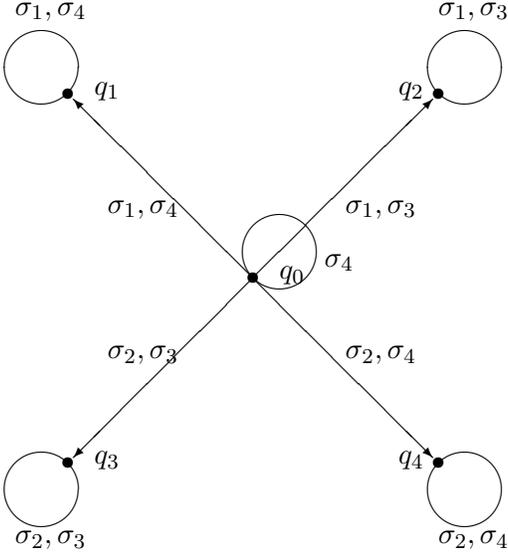

\par
From Fig.~\ref{F:ATMATA}, we know
$Q=\{q_{0},q_{1},q_{2},q_{3},q_{4}\}$ and
$\Sigma=\{\sigma_{1},\sigma_{2},\sigma_{3},\sigma_{4}\}$. It is
easy to compute that: $Q(\sigma_{1})=\{q_{1},q_{2}\}$,
$Q(\sigma_{2})=\{q_{3},q_{4}\}$, $Q(\sigma_{3})=\{q_{2},q_{3}\}$,
$Q(\sigma_{4})=\{q_{0},q_{1},q_{4}\}$.
\par
Diagnosability of the circuit depends on $\Sigma_{o}$ and $T$. Let
the desired partition $T=\{\{q_{0}\},\{q_{1}\},\{q_{3}\}\}$. We
consider the following two  examples for $\Sigma_{o}$.
\par
Let $\Sigma_{o} = \{\sigma_{1},\sigma_{2}\}$, we first use
Algorithm~\ref{A:TQC} to compute $\bigwedge\limits_{\sigma\in
\Sigma_{o}} T_{\sigma}$. In ``Initialization'' section, set $m =
5,n=2,s_{0}=s_{1}=s_{2}=s_{3}=s_{4}=0$. In ``Intersection''recycle,
when step $i=1(\sigma_{1})$, we get $s_{1}=s_{2}=1$; when step
$i=2(\sigma_{2})$, we get $s_{3}=s_{4}=2$. The final result is
$s_{0}=0, s_{1}=s_{2}=1, s_{3}=s_{4}=2$. And then we send the result
to Algorithm~\ref{A:OFD}. By using quicksort algorithm, we get the
result $s_{0}<s_{1}=s_{2}<s_{3}=s_{4}$. In the last statements of
Algorithm~\ref{A:OFD}, we find $s_{1}=s_{2}$. But in the desired
partition $T$, $q_{1}\in\{q_{1}\}$, and
$q_{2}\overline{\in}\{q_{1}\}$, so we get the $OFD=FALSE$ in final.
So the system is not diagnosable with respect to $T$ and
$\Sigma_{o}$.
\par
Let $\Sigma_{o} = \{\sigma_{1},\sigma_{2},\sigma_{3}\}$, we first
use Algorithm~\ref{A:TQC} to compute $\bigwedge\limits_{\sigma\in
\Sigma_{o}} T_{\sigma}$. In ``Initialization'' section, set $m =
5,n=3,s_{0}=s_{1}=s_{2}=s_{3}=s_{4}=0$. In ``Intersection''recycle,
when step $i=1(\sigma_{1})$, we get $s_{1}=s_{2}=1$; when step
$i=2(\sigma_{2})$, we get $s_{3}=s_{4}=2$; when step
$i=3(\sigma_{3})$, we get $s_{2}=5, s_{3}=6$. The final result is
$s_{0}=0, s_{1}=1, s_{2}=5, s_{3}=6, s_{4}=2$. And then we send the
result to Algorithm~\ref{A:OFD}. By using quicksort algorithm, we
get the result $s_{0}<s_{1}<s_{4}<s_{2}<s_{3}$. In the last
statements of Algorithm~\ref{A:OFD}, all the values of
$s_{j}(j=0,1,2,3,4)$ are not equal, and we get the $OFD=TRUE$ in
final. So the system is diagnosable with respect to $T$ and
$\Sigma_{o}$.
\subsection{Lost of  Algorithm~\ref{A:MINOES}}
Let
\[M=\{\sigma_{1},\sigma_{2},\ldots,\sigma_{10}\}.\]
And the cost of $M$ is
\[C(M)=\{27, 23, 20, 15, 10, 9, 7, 5, 4, 1\},\]
$G$ is diagnosable with respect to given $T$ and an element of
$OES(T)$.
\par $OES(T) = \{E_{1} =\{\sigma_{3}, \sigma_{5}, \sigma_{7},
\sigma_{10}\}\}\cup \{E_{2} = \{\sigma_{3}, \sigma_{5},
\sigma_{8}, \sigma_{9}, \sigma_{10}\}\}\cup\{\Sigma\subseteq M:
E_{1}\subseteq\Sigma,or, E_{2}\subseteq\Sigma\}$. Using
Algorithm~\ref{A:MINOES}, we can get the $minOES = E_{2}$, but the
cost of $E_{2}$ is $20+10+5+4+1=40$. The cost of $E_{1}$ is
$20+10+7+1=38$, so the $minOES$ is not the minimal cost of
$OES(T)$.

\subsection{Example of  Algorithm~\ref{A:MMOES}}
Using Algorithm~\ref{A:MMOES}, in ``Initialization'' section, we get
$minOES=\{\sigma_{3}, \sigma_{5}, \sigma_{8}, \sigma_{9},
\sigma_{10}\}$, $lmS= minL(minOES) = 3$, $cmS= C(minOES) = 40$. And
then we get
\[H=\left\{\begin{array}{l}\{\sigma_{1}, \sigma_{5}, \sigma_{10}\},\\
\{\sigma_{1}, \sigma_{6}, \sigma_{9}\},\\
\{\sigma_{1}, \sigma_{6}, \sigma_{10}\},\\
\{\sigma_{1}, \sigma_{7}, \sigma_{8}, \sigma_{10}\},\\
\{\sigma_{1}, \sigma_{7}, \sigma_{9}, \sigma_{10}\},\\
\{\sigma_{1}, \sigma_{8}, \sigma_{9}, \sigma_{10}\},\\
\{\sigma_{2}, \sigma_{4}, \sigma_{10}\},\\
\{\sigma_{2}, \sigma_{5}, \sigma_{7}\},\\
\{\sigma_{2}, \sigma_{5}, \sigma_{8}, \sigma_{10}\},\\
\{\sigma_{2}, \sigma_{5}, \sigma_{9}, \sigma_{10}\},\\
\{\sigma_{2}, \sigma_{6}, \sigma_{7}, \sigma_{10}\},\\
\{\sigma_{2}, \sigma_{6}, \sigma_{8}, \sigma_{10}\},\\
\{\sigma_{2}, \sigma_{6}, \sigma_{9}, \sigma_{10}\},\\
\{\sigma_{2}, \sigma_{7}, \sigma_{8}, \sigma_{9},
\sigma_{10}\},\\\{\sigma_{3}, \sigma_{4}, \sigma_{8}\},\\
\{\sigma_{3}, \sigma_{4}, \sigma_{9}, \sigma_{10}\},\\
\{\sigma_{3}, \sigma_{5}, \sigma_{6}, \sigma_{10}\},\\
\{\sigma_{3}, \sigma_{5}, \sigma_{7}, \sigma_{10}\},\\
\{\sigma_{3}, \sigma_{5}, \sigma_{8}, \sigma_{9}, \sigma_{10}\},\\
\{\sigma_{3}, \sigma_{6}, \sigma_{7}, \sigma_{9}\},\\
\{\sigma_{3}, \sigma_{6}, \sigma_{7}, \sigma_{10}\},\\
\{\sigma_{3}, \sigma_{6}, \sigma_{8}, \sigma_{9}, \sigma_{10}\},\\
\{\sigma_{3}, \sigma_{7}, \sigma_{8}, \sigma_{9}, \sigma_{10}\},\\
\textit{other not empty subset of above set.}\\
\end{array}\right\}\]
\par
In ``Testing diagnosability'' recycle, we find that only two
elements of $H$($E_{1}$ and $E_{2}$) are diagnosable, and
$C(E_{1})<C(E_{2})$. Therefor we get the minimal cost of $OES(T)$ is
$C(E_{1})$.
\par
If we consider the set $NES$. From Algorithm~\ref{A:NES}, we get the
set $NES$. We partition the finite events set $\Sigma$ into disjoint
subsets $NES=\{\sigma_{3}, \sigma_{5},\sigma_{10}\}$ and
$\Sigma\setminus NES=\{\sigma_{1},\sigma_{2},\sigma_{4},\sigma_{6},
\sigma_{7},\sigma_{8}, \sigma_{9}\}$. Now in
Algorithm~\ref{A:MMOES}, we use the set $(\Sigma\setminus NES)$ as
the set $\Sigma$. The computing procedure is as follows: in
``Initialization'' section, we get $minOES=\{\sigma_{8},
\sigma_{9}\}$, $lmS= minL(minOES) = 8$, $cmS= C(minOES) = 9$. And
 then we get that
\[H=\left\{\begin{array}{l}\{\sigma_{6}\},\\\{\sigma_{7}\},\\
\{\sigma_{8}, \sigma_{9}\},\\
\textit{other not empty subset of above set.}\\
\end{array}\right\}\]
\par
In ``Testing diagnosability'' recycle, we find that only two
elements of $H$($\{\sigma_{7}\}$ and $\{\sigma_{8}, \sigma_{9}\}$)
are diagnosable, and $C(\{\sigma_{7}\})<C(\{\sigma_{8},
\sigma_{9}\})$. Hence we get that the minimal cost of $OES(T)$ is
$C(\{\sigma_{7}\})+C(NES)$. The result is the same as that by using
the method above, but the complexity is greatly reduced.

\section{Conclusion}
In terms of some problems in off-line diagnostics~\cite{Lin94}, in
this paper, we present some off-line diagnostic algorithms to
overcome the shortcomings. We give a general method of judging a
system's off-line diagnosability, which is a polynomial-time
algorithm. And we give an algorithm of how to find the minimal set
in all observable event sets. Of course, another  issue worthy of
further consideration is the on-line diagnostic algorithms of the
minimal cost in DESs. We would like to consider it in subsequent
work.



\end{document}